%% file: main_arxiv.tex
\documentclass[10pt,twocolumn,letterpaper]{article}

\usepackage[pagenumbers]{cvpr} 

\input{preamble}

\input{math_commands.tex}

\usepackage[utf8]{inputenc} 
\usepackage[T1]{fontenc}    
\usepackage{url}            
\usepackage{booktabs}       
\usepackage{amsfonts}       
\usepackage{nicefrac}       
\usepackage{microtype}      
\usepackage{xcolor}         
\usepackage{enumitem}
\usepackage{tikz}
\usepackage{pgfplots}
\usepackage{xspace}
\pgfplotsset{compat=1.18}
\usepackage[most]{tcolorbox}
\usepackage{tabularx}
\usepackage{graphicx}
\usepackage{subcaption}
\usepackage{adjustbox}
\usepackage{multirow}
\usepackage{wrapfig}
\usepackage{placeins}

\usepackage{colortbl}
\usepackage{pifont}
\usepackage{makecell}
\usepackage{arydshln}
\usepackage{amssymb}

\definecolor{draftgray}{gray}{0.48}

\definecolor{bestgreen}{HTML}{DDEBD5}
\newcommand{\best}[1]{\cellcolor{bestgreen}\underline{#1}}

\definecolor{cvprblue}{rgb}{0.21,0.49,0.74}
\usepackage[pagebackref,breaklinks,colorlinks,allcolors=cvprblue]{hyperref}

\def\confName{CVPR}
\def\confYear{2027}

\title{Recursive Self-Improvement in Unified Multimodal Models}

\author{
Huijuan Wang$^{1,*}$ \quad
Chufan Shi$^{1,*}$ \quad
Cheng Yang$^{2,*}$ \quad
Yaokang Wu$^{3}$ \\
Taylor Berg-Kirkpatrick$^{2}$ \quad
Xuezhe Ma$^{1}$ \\[0.6em]
$^{1}$University of Southern California \quad
$^{2}$University of California San Diego \quad
$^{3}$Carnegie Mellon University \\
{\tt\small huijuanw@usc.edu}
}

\begin{document}
\maketitle
{\renewcommand{\thefootnote}{\fnsymbol{footnote}}\footnotetext[1]{Equal contribution.}}
\input{sec/0_abstract}    
\input{sec/1_introduction}
\input{sec/2_method}
\input{sec/3_benchmark}
\input{sec/4_experiment}

\input{sec/5_related_work}
\input{sec/6_conclusion}
{
    \small
    \bibliographystyle{unsrtnat}
    \bibliography{main}
}
\clearpage
\appendix
\crefalias{section}{appendix}
\crefalias{subsection}{subappendix}
\input{appendix/00_implementation_details}

\input{appendix/01_request_sets}

\input{appendix/02_evaluation_protocol}

\input{appendix/04_additional_results}

\input{appendix/05_data_distribution_changes}


\end{document}

%% file: preamble.tex
\newcommand{\blue}[1]{{\color{black}#1}}
\newcommand{\rev}[1]{{\color{black}#1}}

\makeatletter
\@ifpackageloaded{lineno}{%
  \IfFormatAtLeastTF{2025-06-01}{%
    \RemoveFromHook{build/column/before}[lineno]%
    \AddToHook{build/column/after}{%
      \setbox\@outputbox\vbox{%
        \boxmaxdepth\@maxdepth
        \protected@write\@auxout{}{%
          \string\def\string\@LN@column{\if@firstcolumn1\else2\fi}}%
        \box\@outputbox}}%
  }{}%
}{}
\makeatother

%% file: math_commands.tex
\usepackage{amsmath,amsfonts,bm}

\def\eqref#1{equation~\ref{#1}}

\def\1{\bm{1}}

\DeclareMathAlphabet{\mathsfit}{\encodingdefault}{\sfdefault}{m}{sl}
\SetMathAlphabet{\mathsfit}{bold}{\encodingdefault}{\sfdefault}{bx}{n}



%% file: sec/0_abstract.tex
\begin{abstract}
Unified multimodal models (UMMs) understand and generate both text and images, which lets a model produce its own training data. Existing self-improvement in UMMs keeps supervision on the visual side, where image understanding judges image generation. We propose recursive cross-capability self-improvement (RSI), a training loop in which the text and visual abilities of a UMM supply training data for one another. In each round, the model \blue{acts as a Painter that} generates images and \blue{an Inspector that} reads them to find where it falls short. \blue{As a Coder, it} then writes programs aimed at these shortcomings, and \blue{an Executor runs them and verifies} every result against its specification. Verified renders train \blue{the Painter}, while labeled renders and the model's own correct programs train \blue{the Inspector and the Coder}. Program execution thus acts as a source of truth outside the model, so errors do not accumulate across rounds. We study RSI on charts and build BasicChartBench, \blue{which checks every element of a generated chart, from its structure and text to how values are bound to categories, with atomic questions}. \blue{On BasicChartBench, three rounds of RSI raise question accuracy from 26.6\% to 31.7\%, while continued training reaches only 27.3\%. Verified construction carries most of the gain, and targeting the model's failures adds 0.6\% on BasicChartBench and 1.7\% on StructT2IBench after two rounds.} \blue{Along the way, the Inspector's accuracy on a fixed chart understanding set rises from 72.7\% to 94.2\% in question answering and from 17.0\% to 84.1\% in verification.}
\end{abstract}

%% file: sec/1_introduction.tex
\section{Introduction}
\label{sec:introduction}

Progress in foundation models has largely been driven by data that humans collect, filter, and annotate. Unified multimodal models (UMMs) suggest a different path. A UMM understands and generates both text and images within a single model~\citep{team2024chameleon,zhou2025transfusion,deng2025bagel}. It could therefore produce its own training data, with one ability supervising another. Self-improvement of this kind has already advanced the reasoning of language models~\citep{zelikman2022star,yuan2024selfrewarding}. In UMMs, image understanding has recently begun to reward or filter the outputs of image generation~\citep{mao2025unirl,jin2025srum,han2026unicorn}. This supervision stays on the visual side, and the text abilities of the model take no part in it.

\begin{figure*}[t]
\centering
\includegraphics[width=\textwidth]{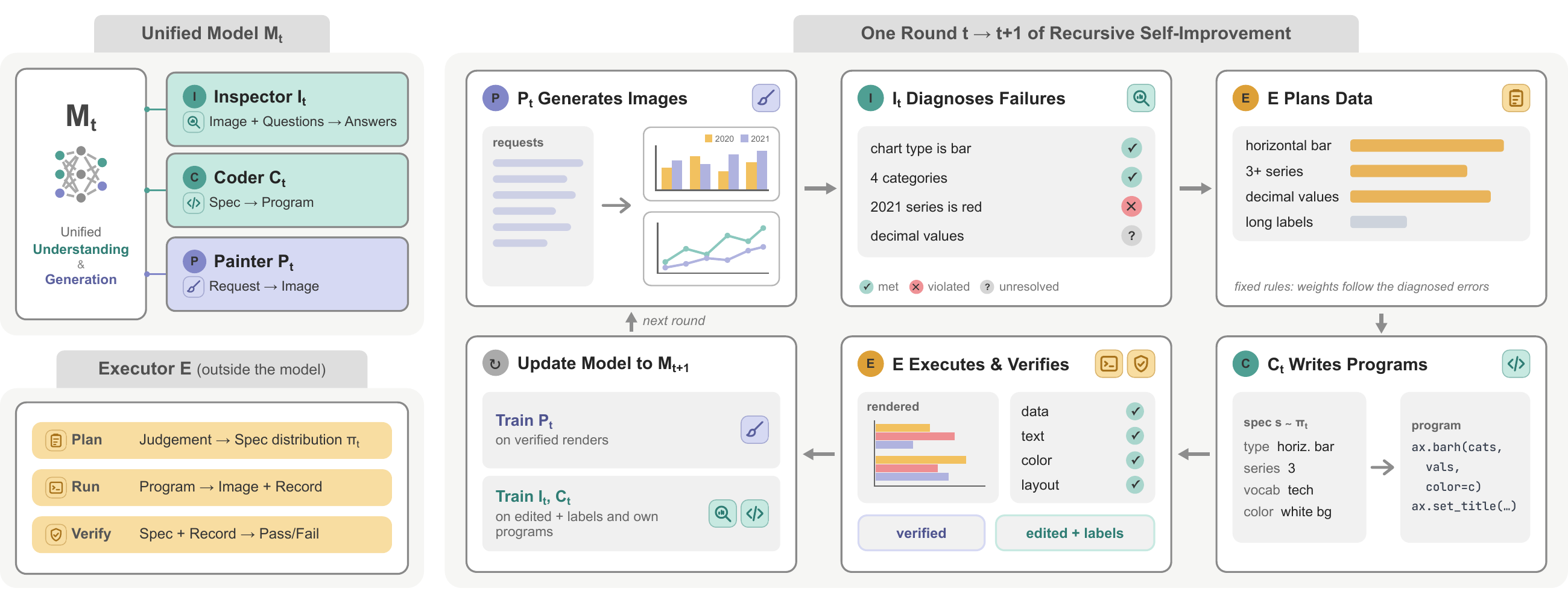}
\caption{One round of RSI. \blue{The Inspector reads the Painter's generations, and fixed rules turn this diagnosis into a data plan. The Coder writes programs for the plan, and the Executor runs and verifies them, supplying verified images and exact labels.} The updated model then changes the errors \blue{of the Painter}, the \blue{judgments of the Inspector}, and the \blue{programs of the Coder} in the following round.}
\label{fig:recursive_feedback}
\end{figure*}

We ask whether the text and visual abilities of a UMM can improve one another in a closed loop. \blue{As a Painter, the model} shows what it cannot yet draw. \blue{As an Inspector, it} finds what is wrong, and \blue{as a Coder, it} describes what should be drawn instead. Each ability turns what it finds into training data for the others. Round after round, the data follow the model's changing weaknesses~\citep{zhao2024evolvedirector,khan2025dataenvgym}, and each gain strengthens the abilities that build the next round. The risk is that the loop feeds on its own outputs. Without an independent check, the errors of one round are learned in the next, and they compound over rounds~\citep{shumailov2024collapse}.

The loop therefore needs a source of truth outside the model. For structured visual content, programs provide one. Infographics, diagrams, tables, and documents can all be written as programs~\citep{yang2025cosyn,zhuo2026factuality}. Running a program yields an image and an exact record of what it contains, so correctness comes from execution rather than from the model's own judgment. Building on this idea, we propose \emph{recursive cross-capability self-improvement} (RSI). In each round, the model \blue{acts as a Painter that} generates images and \blue{an Inspector that} reads them to find where it falls short. \blue{As a Coder, it} then writes programs aimed at these shortcomings, and \blue{an Executor runs them and verifies} every result. Verified renders train \blue{the Painter}. \blue{Edited renders with exact labels train the Inspector, and the model's own correct programs and programs for the charts it still draws wrongly train the Coder} (\Cref{fig:recursive_feedback}). The model decides what to learn, and execution ensures that its data are correct.

We study RSI on charts, a representative form of structured visual content. A chart must be exact in its text, values, and structure, and open models remain weak at this. \blue{Existing benchmarks~\citep{zhuo2026factuality,li2026bizgeneval} give an overall score, but they do not show which details of a chart a model gets wrong. We therefore build BasicChartBench, which derives from each specification one atomic question for every requirement it states, across nine chart types and a range of panel counts, category counts, and label lengths.}

\blue{Over three rounds, generation improves in every round. On BasicChartBench, question accuracy rises from 26.6\% to 31.7\%, while continued training on the original data reaches only 27.3\%. \blue{On the chart subset of StructT2IBench, accuracy rises from 11.0\% to 26.7\%.}} Verified construction carries most of this gain, and building data where the model fails adds to it. The text and visual abilities also improve along the way. After one round, \blue{the Inspector} catches far more of the errors planted in rendered charts. We close by analyzing in detail where the loop does not yet help.

Our contributions are as follows:
\begin{itemize}[itemsep=2pt, parsep=0pt, topsep=0pt, partopsep=0pt, leftmargin=*, beginpenalty=10000]
 \item We introduce the idea of letting the text and visual abilities of a UMM supervise one another in a closed loop, with program execution as an independent check.
 \item We realize this idea as RSI and apply it to charts. \blue{To evaluate generation detail by detail}, we build BasicChartBench, with \blue{840} charts and \blue{52{,}169} atomic questions \blue{covering structure, text, and binding}.
 \item \blue{Over three rounds, RSI raises question accuracy on BasicChartBench from 26.6\% to 31.7\%, while continued training reaches only 27.3\%. After two rounds, targeting the model's failures adds 0.6\% on BasicChartBench and 1.7\% on StructT2IBench over spreading the same data evenly.} \blue{Along the way, the Inspector's accuracy on a fixed chart understanding set rises from 72.7\% to 94.2\% in question answering and from 17.0\% to 84.1\% in verification.}
\end{itemize}

%% file: sec/2_method.tex
\section{Method}
\label{sec:method}

RSI runs in rounds that follow the loop in \Cref{fig:recursive_feedback}. \Cref{sec:setup} sets up notation, and \Cref{sec:executable_experience} shows how execution supplies a source of truth. \Crefrange{sec:native_diagnosis}{sec:recursive_update} describe the three steps of a round: diagnosing the model's own generations, building verified data, and updating the model.

\subsection{Setup}
\label{sec:setup}

Let $M_t$ denote the UMM after round $t$, with $M_0$ the starting model. We use three of its abilities\blue{, each acting as a role}. \blue{The Painter $P_t$ draws} an image from a text request. \blue{The Inspector $I_t$} answers questions about an image. \blue{The Coder} $C_t$, \blue{which uses} text generation, turns a specification into a program. The model keeps separate expert parameters for understanding and generation. \blue{$I_t$} and $C_t$ share the understanding parameters, and \blue{$P_t$} uses the generation parameters. \blue{A non-learned Executor $\mathcal E$, outside the model, runs programs and records what they draw, verifies these records against the specification, and labels questions with what the images show.}
\looseness=-1

A specification $s$ fixes the content of one chart. We describe it along predefined capability dimensions, such as chart type, number of categories, number of series, color similarity, value labels, number format, label length, and vocabulary. Most dimensions have ordered levels of difficulty, such as one, two to three, or four to six series. A request $q(s)$ states every requirement of $s$ in natural language, in one of several wordings. Atomic questions derived from $s$ check the requirements one by one. Each question belongs to one of three groups: structure, text, or binding (\Cref{sec:benchmark}).

\subsection{Execution as a Source of Truth}
\label{sec:executable_experience}

For a specification $s$, the \blue{Coder} produces a program $p=C_t(s)$. Running it gives an image \blue{$x$} and a record \blue{$z$} of what was drawn,
\begin{equation}
 (\blue{x},z)=\mathcal E(p).
\end{equation}
\blue{The Executor then compares} the record $z$ with $s$. It checks the data, text, colors, axes, layout, and text readability of the chart. \blue{Only passing pairs are kept. Execution thus verifies every target. Correctness is decided by execution rather than by the model's own judgment. This keeps the model's own errors out of its training data and prevents them from accumulating across rounds. The accepted pairs $(q(s),x)$ become targets for generation training. They directly improve the Painter's image generation ability.}

\blue{The same records also supply training data for the Inspector. The Inspector learns to judge both images that meet a request and images that do not. Correct images come directly from rendering the verified programs, so every requirement of $s$ is met. Incorrect images need controlled errors. We change one property of $s$, such as swapping two colors or reordering categories, and render the chart again. The Executor then labels each atomic question of $s$ with what the rendered image shows, not with what the request asks for. For example, suppose the request asks for a red sales series and a blue cost series. After the two colors are swapped, the image draws the sales series in blue, so the question on its color is labeled as not met. The Coder shares the understanding parameters with the Inspector and is trained on the model's own verified programs paired with their specifications. Together, these examples form the understanding data $\mathcal D_t^U$.}
\looseness=-1

\subsection{Diagnosing the Model's Own Generations}
\label{sec:native_diagnosis}

\blue{Each round begins with locating the Painter's current weaknesses on a fixed set of diagnostic requests. For each chart type, the set contains 48 requests, each with three to five categories, such as the bars along the x-axis or the slices of a pie. These requests vary the number of series, the series names, the value magnitude, the number format, and the request wording in a balanced design, so any two values of an attribute are compared on otherwise matched requests. The wordings are those used in training. When a new vocabulary opens, we add a half-sized set with the same design. None of these requests overlaps with the test set.} For each request $q(s)$, the current \blue{Painter} produces an image $\blue{\widetilde x}=\blue{P_t}(q(s))$. \blue{Unlike a program render, this image comes with no record of what was drawn. The Inspector therefore checks its content by answering the atomic questions $Q(s)$ of $s$,}
\begin{equation}
 \widehat a=\blue{I_t}(\blue{\widetilde x}\blue{,Q(s)}),\qquad d=\operatorname{Compare}(\widehat a,s),
\end{equation}
where $d$ marks each requirement as met, violated, or unresolved. \blue{A requirement is unresolved when the answer is empty or cannot be compared with $s$. The raw generation accuracy is the share of met requirements among the resolved ones, and we report the unresolved share separately. We compute it for each capability dimension, for each request wording, and for each attribute value, that is, each setting of a property of $s$, such as the value magnitude, the number format, or the vocabulary.}

\blue{Three further signals check the diagnosis in every round. First, programs render each diagnostic request correctly. These renders satisfy every requirement by construction, so any requirement that the Inspector marks as not met on them is an Inspector error. We measure this false-rejection rate for each pair of chart type and capability dimension and add it back to the generation accuracy of that pair. Second, held-out edited renders test whether the Inspector reports what the image shows or what the request asks. Its error rate on these renders sets how much Inspector training data each pair receives. Third, optical character recognition (OCR)\blue{~\citep{cui2025paddleocr30technicalreport}} checks the rendered text independently of the model. Its score on held-out topics decides when the training text includes more unseen real words, and when longer titles and new number formats are introduced.}

\subsection{From Diagnosis to Verified Data}
\label{sec:targeted_construction}

\blue{Fixed rules turn the diagnosis into the data of the next round. The rules are deterministic and fixed before the round in which they are used, so no proportion is adjusted by hand between rounds. Wherever a budget is divided among items, the rules use the same form: 20\% is spread evenly to keep learned skills practiced, and the rest follows the measured error rates, capped at three times the mean so that one weakness cannot take over a round.}

\blue{\textbf{Generation plan.} The plan is a sampling distribution $\pi_t$ over specifications. The budget is divided among the pairs of chart type and capability dimension by the Painter's error rate on each pair. Each pair has a difficulty level that moves up when its accuracy reaches 80\% and down when it falls below 60\%, and most of its specifications are drawn at the current level. Within a pair, attribute values that remain weak receive a larger share, and the size of the increase depends on whether the value was rare in earlier training data or is hard despite frequent training. A curriculum widens the space of specifications, for example with new number formats, longer titles, more unseen words, and new chart types, once fixed gates on the diagnosis are met.}

\blue{\textbf{Understanding plan.} The shares of the task types are fixed, and the diagnosis decides which chart questions are asked. Edited renders are built anew in every round. Each type of edit is weighted by the product of two rates: how often the Painter makes the matching error, and how often the Inspector fails to see it. The Inspector is therefore trained most on the errors that the Painter makes and that it currently misses. Before the plans are derived, the new Inspector must not be worse than the previous one on any group of questions.}

The \blue{Coder} produces two candidate programs for each specification drawn from $\pi_t$. We keep the first verified candidate and fill each pair up to its share,
\begin{equation}
 \mathcal D_t^G=\operatorname{Verify}\!\left(
 \left\{\bigl(q(s),\blue{x}\bigr)\mid s\sim\pi_t,\ (\blue{x},z)=\mathcal E(C_t(s))\right\}
 \right).
\end{equation}
Verification also shows where the \blue{Coder} is weak, through its pass rate on each attribute value. Values with a low pass rate receive extra specifications in the next round, up to 20\% of the round. Their correct programs are also sampled up to three times as often for the understanding data.

\subsection{Updates and Recursion}
\label{sec:recursive_update}

\blue{A single update trains all parameters of $M_t$ on both sets of data,}
\looseness=-1
\begin{equation}
 \blue{M_{t+1}=\operatorname{Learn}\bigl(M_t;\ \mathcal D_t^G\cup\mathcal D_t^U\bigr).}
\end{equation}
\blue{It optimizes the image generation loss on $\mathcal D_t^G$ and the text loss on $\mathcal D_t^U$ together, so the Painter, the Inspector, and the Coder all change in the same update. The Painter reads the request through the understanding parameters. Under a joint update, the generation loss therefore also adapts how the model reads chart requests, and the generation parameters are trained against the same understanding parameters that they are used with afterwards. The alternative is to train the two sets of parameters separately, each on its own data with the other frozen, and to merge them afterwards. This cuts the flow of information between understanding and generation, and we compare the two in \Cref{sec:ablations}.}

In the next round, \blue{$P_{t+1}$} makes new errors, \blue{$I_{t+1}$} reads them, and $C_{t+1}$ writes the new examples. The curriculum gates open new parts of the specification space once the diagnosis shows that the earlier parts are learned, so RSI sets its own training space from its diagnosis. At inference time, the \blue{Painter} draws images directly, and programs are used only to build training data.

%% file: sec/3_benchmark.tex
\section{BasicChartBench}
\label{sec:benchmark}

\begin{figure}[t]
\centering
\includegraphics[width=\linewidth]{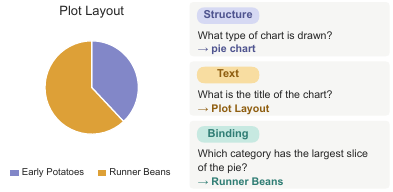}
\caption{\blue{An example in BasicChartBench: a request, its reference render, and one atomic question per group.}}
\label{fig:bcb_example}
\end{figure}

\blue{A chart is correct only if every element is correct, from its type and layout to each label, number, and color. Existing chart generation benchmarks~\citep{zhuo2026factuality} report how often a model meets a request, but their scores do not show which of these elements it gets wrong. We therefore build BasicChartBench, which checks every requirement of a request on its own (\Cref{fig:bcb_example}). A program derives the questions from the full specification and groups them by kind, so a score can be traced to the structure, the text, or the binding of values to categories.}
\looseness=-1

\begin{figure}[t]
\centering
\includegraphics[width=\linewidth]{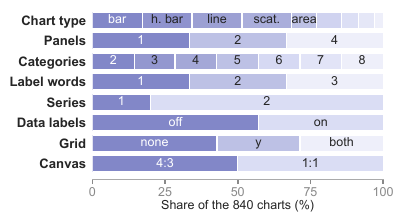}
\caption{\blue{How the 840 charts of BasicChartBench are spread over its design factors. The four unlabeled chart types are histograms, pies, rings, and heatmaps, in this order.}}
\label{fig:bcb_distribution}
\end{figure}

\begin{figure*}[t]
\centering
\begin{minipage}[t]{\columnwidth}
\centering
\includegraphics[width=\linewidth]{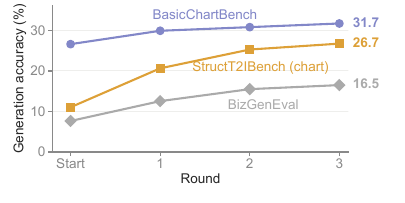}
\caption{\blue{Accuracy of the Painter after each round of RSI on BasicChartBench, the chart subset of StructT2IBench, and BizGenEval.}}
\label{fig:round1}
\end{minipage}\hfill
\begin{minipage}[t]{\columnwidth}
\centering
\includegraphics[width=\linewidth]{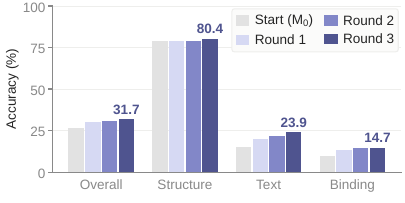}
\caption{\blue{Accuracy of the Painter on BasicChartBench after each round, over all questions and for each question group.}}
\label{fig:round_final}
\end{minipage}
\end{figure*}

\blue{\textbf{Charts.} BasicChartBench has 840 charts of nine types: vertical bars, horizontal bars, lines, scatter plots, stacked areas, histograms, pies, rings, and heatmaps. Its design factors are varied independently of one another (\Cref{fig:bcb_distribution}). A chart has one, two, or four panels of the same type, and two to eight categories. Category labels have one, two, or three words. The remaining factors are the number of series, value labels on the marks, the grid, and the aspect ratio of the canvas. Values and colors are drawn afresh for every chart, and all values are integers. BasicChartBench is never used for diagnosis or data selection.}

\blue{\textbf{Questions.} Each chart comes with atomic questions, 52{,}169 in total and 62 per chart on average. Each question checks one requirement and asks for the answer directly, such as a title, a count, a list of labels, or the value of one mark, rather than for a yes or a no. As in \Cref{fig:bcb_example}, the questions form three groups. Structure questions (12{,}011) ask about the chart type, the number and layout of the panels, whether a legend and data labels are shown and where the legend sits, the grid, the orientation of the axes, and the geometry and spacing of the marks. Text questions (10{,}776) ask for the title, the axis titles, the panel letters, the category labels, the legend entries, and the number format exactly as requested. Binding questions (29{,}382) ask which value, color, marker, or line style belongs to which category or series, in what order the series appear, what the data labels show, and how the tick labels are formatted. For a finer view, the questions are also divided into 25 categories, of which reading a value from a mark is the largest, with 14{,}421 questions (\Cref{sec:appendix_sets}).}

\blue{\textbf{Scoring.} An external judge, Qwen3.6-35B-A3B~\citep{qwen36_35b_a3b}, answers every question from the Painter's image alone, and deterministic rules compare each answer with the specification. Text must match exactly, and a value read from a mark counts as correct within 5 units or 10\% of the target. The judge answers N/A when the image does not show what a question asks for, and such an answer counts as wrong. A model's score is its question accuracy, the share of questions answered correctly, and we also report the mean of the per-chart accuracies. On the reference renders, the judge reaches 98.3\%, which bounds the error that the judge itself adds to a score.}
\looseness=-1

%% file: sec/4_experiment.tex
\section{Experiments}
\label{sec:experiments}

We ask four questions in turn. Does the loop keep improving \blue{the Painter} (\Cref{sec:main_results})? Which of its parts drive the gain (\Cref{sec:ablations})? How do the text and visual abilities of the model change along the way (\Cref{sec:analysis})? And where does the loop still fall short (\Cref{sec:shortfalls})?

\subsection{Experimental Setting}
\label{sec:setting}

We build on BAGEL-7B-MoT~\citep{deng2025bagel}. The starting model $M_0$ comes from a curriculum that goes from easy to hard. The model first learns to write plotting programs. Generation then learns basic chart types with short text, followed by a larger set of programmatically generated charts. Finally, the understanding parameters learn to answer atomic questions about program-rendered charts (\Cref{sec:appendix_start}). The tasks cover seven chart types that verification can check: vertical bars, horizontal bars, lines, scatter plots, areas, pies, and rings. A diagnostic set of 336 requests in the vocabulary of the first curriculum stage drives the data plan in every round. It varies the series count, value size, and request wording. Every round uses the same training budget (\Cref{sec:appendix}).

\textbf{Controls.} Each control removes one part of RSI. \emph{Continued training} trains the \blue{Painter} from $M_0$ for the same number of steps on the original data mixture of $M_0$ and tests whether the gain comes only from more training. The \emph{fixed distribution} \blue{does not use the diagnosis. It spreads the generation data evenly over chart types and capability dimensions and the \blue{Inspector} data evenly over the types of edit, with the same curriculum as RSI.} \emph{Single diagnosis} \blue{diagnoses only $M_0$. It keeps the resulting plan and curriculum in every round, and $M_0$ writes all of its programs, so its first round is that of RSI. Both controls run for two rounds.}

\subsection{Does the Loop Keep Improving?}
\label{sec:main_results}

%

\blue{\Cref{fig:round1} shows the Painter after each round. On BasicChartBench, question accuracy rises in every round, from 26.6\% at the start to 29.9\%, 30.8\%, and 31.7\% in the three rounds, and the mean of the per-chart accuracies rises from 28.8\% to 35.4\%. The score rises in every round on the two existing benchmarks as well, from 11.0\% to 26.7\% on the chart subset of StructT2IBench and from 7.6\% to 16.5\% on BizGenEval. BasicChartBench remains far from solved, however, as the judge reaches 98.3\% on its reference renders.}

\blue{Most of the gain comes from text and binding (\Cref{fig:round_final}). Over three rounds, text accuracy rises from 14.9\% to 23.9\% and binding from 9.6\% to 14.7\%, while structure, which is already high at the start, moves from 78.7\% to 80.4\%.}

\blue{The categories show which details improve (\Cref{fig:bcb_categories}). The largest gains are in the legend position (20.2\% to 54.8\%), the title (0.1\% to 26.9\%), the number format (16.3\% to 42.4\%), the grid (47.2\% to 70.2\%), and the bin edges of histograms (4.8\% to 26.8\%). Other details barely move. Reading a value from a mark, the largest category, stays at 3.0\%, although the judge answers 96.5\% of these questions on the reference renders, and the axis titles and the category list, which must be exact in every word, stay near 5\%. A few categories fall, most of all \blue{the heatmap row count (33.9\% to 21.4\%) and} the panel letters (80.9\% to 73.4\%). \Cref{tab:bcb_categories} lists every category after each round.}

\subsection{Which Parts Drive the Gain?}
\label{sec:ablations}

\begin{figure*}[t]
\centering
\includegraphics[width=\textwidth]{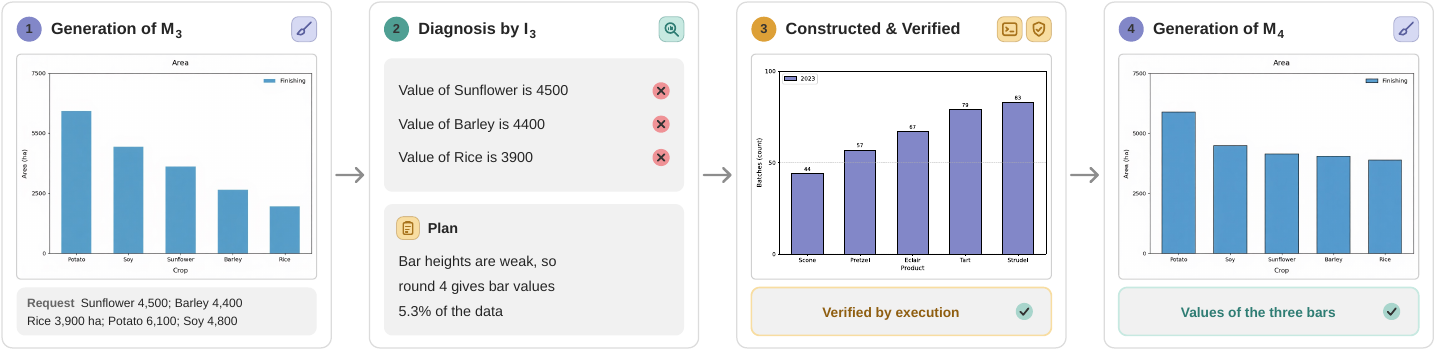}
\caption{One round of RSI on a diagnostic request. (1) After round 3, the \blue{Painter} draws three bars far too low. (2) The \blue{Inspector} \blue{$I_3$} marks exactly these three values as violated, and the plan of round 4 gives this kind of requirement more data. (3) A verified example of the same kind built in round 4. (4) After round 4, the same request is drawn with the three values correct and all text exact.}
\label{fig:qualitative}
\end{figure*}

\textbf{Diagnosis decides which examples to build.} The fixed distribution keeps everything of RSI except the diagnosis. \blue{RSI is ahead of it after both rounds on both benchmarks (\Cref{tab:rounds_levels}), and its lead on StructT2IBench grows from 0.2\% to 1.7\%. Continued training gains only 0.7\% on BasicChartBench, so most of the gain of both strategies comes from the verified charts and not from more training steps.}

\blue{The question groups show where the diagnosis helps. Text improves just as much without it, to 21.9\% for both after the second round. The difference lies in structure and binding. Under the fixed distribution, structure falls below its starting value after the first round (76.8\% against 78.7\%), while RSI stays at 79.0\%, and binding after the second round is 13.6\% against 14.4\% for RSI. By category, RSI leads most in the legend position, the grid, the markers, and the bin edges, by 5.5\% to 7.7\% after the second round. The lead is not uniform, as the fixed distribution is slightly ahead in the panel count and the layout after the second round. RSI is ahead at all three text lengths, by 0.2\% to 1.1\%, while charts with seven or eight categories stay hard for every model, between 24\% and 30\%. Choosing examples from the errors in the model's own outputs therefore adds to what verified charts alone provide, mainly in the details of a chart other than its text.} \Cref{fig:qualitative} shows one such case in detail.

\input{tables/rounds_levels}

\textbf{Recursion follows the changing model.} As the model learns, its errors change\blue{, and single diagnosis does not follow them. On BasicChartBench, the two are within 0.3\% after the second round, with single diagnosis slightly ahead (\Cref{tab:rounds_levels}). It is higher in structure and in text, by about 1\%, and in the legend position, the grid, and the panel letters, while RSI is higher in binding and in the bin edges. On StructT2IBench and BizGenEval, diagnosing the model again brings an improvement. After the second round, RSI leads single diagnosis by 1.2\% on the chart subset of StructT2IBench and by 2.2\% on BizGenEval, with 15.5\% against 13.3\%.} Recursion also changes \blue{the Inspector and the Coder, which we examine} in the next section.

\blue{\textbf{Joint updates carry the data into generation.} RSI updates all parameters of the model at once (\Cref{sec:recursive_update}). To test this choice, we train the first round again on the same data but update the generation and the understanding parameters separately and merge them afterwards (\Cref{tab:joint_update}). With separate updates, the Painter improves far less. On the chart subset of StructT2IBench, it reaches 11.7\% against 20.6\% with the joint update, from 11.0\% at the start. The Painter reads each request through the understanding parameters, and only the joint update trains the generation parameters together with the understanding parameters they are used with afterwards. Separate updates keep general understanding slightly better, and the Coder is about equal. Weighing the three abilities together, we use the joint update.}

\input{tables/joint_update}

\subsection{How Do the Text and Visual Abilities Change?}
\label{sec:diagnosis_experiment}\label{sec:analysis}

\begin{figure}[t]
\centering
\begin{subfigure}{0.325\linewidth}
\centering
\includegraphics[width=\linewidth]{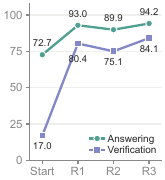}
\caption{\blue{Understanding set}}
\label{fig:abilities_reading}
\end{subfigure}\hfill
\begin{subfigure}{0.325\linewidth}
\centering
\includegraphics[width=\linewidth]{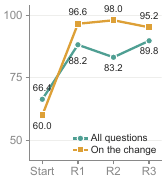}
\caption{\blue{Controlled errors}}
\label{fig:abilities_edits}
\end{subfigure}\hfill
\begin{subfigure}{0.325\linewidth}
\centering
\includegraphics[width=\linewidth]{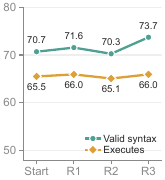}
\caption{\blue{Coding}}
\label{fig:abilities_coding}
\end{subfigure}
\caption{\blue{The understanding parameters after each round of RSI. All values are percentages. (a) Accuracy of the Inspector on the fixed chart understanding set, in question answering and in verification. (b) Accuracy of the Inspector on renders with controlled errors, over all questions and on the questions about the changed property. (c) Share of the Coder's programs on 232 programming tasks that have valid syntax and that execute. Start is the model $M_0$, and the y-axes differ across panels.}}
\label{fig:abilities}
\end{figure}

\blue{The Coder and the Inspector} share the understanding parameters, so every understanding update changes both of them. \blue{We follow the Inspector on a fixed chart understanding set and on renders with controlled errors, the Coder on a fixed set of programming tasks, and the Inspector again on the Painter's own generations.}

\blue{\textbf{The Inspector on a fixed understanding set.} The set has 342 charts with 3{,}939 question-answering questions and 1{,}026 verification questions (\Cref{fig:abilities_reading}). One round raises question answering from 72.7\% to 93.0\% and verification from 17.0\% to 80.4\%. The gain is not monotonic. After the second round, question answering falls to 89.9\% and verification to 75.1\%. The loss is in the yes-or-no questions, which fall from 95.1\% to 88.3\%, while the open questions stay at 91.3\% and 91.6\%. The third round recovers the loss and gives the best values, 94.2\% and 84.1\%. Building the Inspector data from the diagnosis brings no measurable benefit on this set. After the first round, the fixed distribution is close to RSI, with 92.2\% against 93.0\% in question answering and 78.6\% against 80.4\% in verification. After the second round, both controls are ahead of RSI, with 93.2\% and 84.3\% for the fixed distribution and 92.7\% and 82.7\% for single diagnosis, because neither shows the drop of RSI's second round.}

\blue{\textbf{The Inspector on renders with controlled errors.} These renders test whether the Inspector reports what an image does not show. The probe has 1{,}248 renders, with and without a controlled error, and 17{,}946 questions, of which 1{,}030 ask directly about the property that the error changes (\Cref{fig:abilities_edits}). Over all questions, accuracy rises from 66.4\% to 88.2\% after one round, falls to 83.2\% after the second, and reaches 89.8\% after the third, as on the understanding set. The drop of the second round does not come from the errors. On the questions about the changed property, accuracy rises from 60.0\% to 96.6\% after one round and stays between 95.2\% and 98.0\% afterwards, and the questions on renders with an error fall no more in the second round than those on renders without one (to 83.4\% and 83.1\%). The Inspector thus keeps reporting the errors, and what it loses in the second round is general reading accuracy.}

\begin{figure}[t]
\centering
\includegraphics[width=\linewidth]{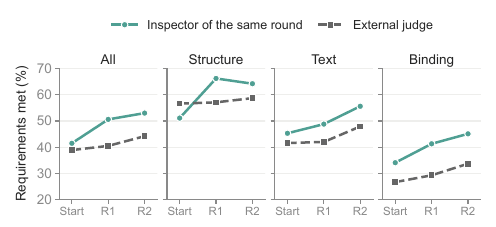}
\caption{\blue{Share of requirements met in the Painter's own generations after each round of RSI, as read by the Inspector of the same round and by the external judge, on a probe of 555 charts with 9{,}298 questions. Start is the model $M_0$.}}
\label{fig:self_vs_judge}
\end{figure}

\blue{\textbf{The Coder.} We measure the Coder on 232 tasks that ask for a plotting program, with greedy decoding and a limit of 16{,}384 tokens (\Cref{fig:abilities_coding}). The loop keeps this ability but improves it little. The share of programs with valid syntax is 70.7\% at the start and 71.6\%, 70.3\%, and 73.7\% after the three rounds, and the share that executes stays between 65.1\% and 66.0\%. Most failures are programs that reach the token limit, which happens on 29.3\% of the tasks at the start and on 26.3\% after round 3. The two controls lose part of this ability. After two rounds, 60.8\% of the programs execute under the fixed distribution and 61.2\% under single diagnosis, against 65.1\% for RSI.}

\textbf{\blue{The Inspector on the model's own generations.}} Program renders cannot show whether the \blue{Inspector} also reads the model's own generations, which are the input of every diagnosis. \blue{We compare the Inspector with the external judge on a probe of 555 charts with 9{,}298 questions, in which the images of each Painter are read both by the Inspector of the same round and by the judge. Both find more requirements met after each round, but the Inspector finds a larger gain, from 41.5\% to 53.0\% over two rounds against 38.9\% to 44.2\% for the judge. The two agree on about 73\% of the questions in every round. The questions that only the Inspector accepts grow from 1{,}392 at the start to 1{,}659 after round 2, and those that only the judge accepts fall from 1{,}149 to 833, so the updated Inspectors are more lenient than the judge \blue{(\Cref{fig:self_vs_judge})}. \blue{Structure shows this most clearly. The Inspector finds structure met on 51.1\% of the questions at the start and on 66.2\% after round 1, while the judge stays between 56.6\% and 58.7\%. Most of this jump comes from the chart type, which the Inspector accepts on 45.8\% of the questions at the start and on 90.8\% after round 1, against 68.5\% and 71.8\% for the judge.} About 12\% of the judge's answers on this probe are cut off at its limit of 192 tokens and count as not met, so its rates are somewhat low.}

\subsection{Where Does the Loop Fall Short?}
\label{sec:shortfalls}\label{sec:public_benchmarks}

\blue{\textbf{General abilities.} Almost all of the data that RSI builds are charts, so we also follow the general abilities of the model after each round (\Cref{tab:public}). General generation is kept. GenEval~\citep{ghosh2023geneval} is 82.9 at the start and between 84.4 and 85.2 after every round. General understanding changes in two directions. MM-Vet~\citep{yu2023mm} rises in every round, from 50.3 to 57.1, and MMBench~\citep{liu2024mmbench} rises slightly, from 82.7 to 83.9, while MME~\citep{fu2026mme} stays within 0.3\% of its starting score. MMMU~\citep{yue2024mmmu} instead falls in every round, from 52.1 to 48.9. The gain on MM-Vet is broad. All six of its capabilities rise, including OCR (59.4 to 66.1), spatial awareness (59.2 to 65.9), and math (43.5 to 58.5), which the atomic questions of the \blue{Inspector} data exercise directly, but also knowledge (30.0 to 41.4) and language generation (29.9 to 43.1), which they do not. MM-Vet is the only one of these benchmarks with free-form answers, and the median length of the model's answers on it grows from 5 words at the start to 51 after round 3. Part of the gain may therefore come from fuller answers rather than from better perception alone, and MMBench and MME, which ask for a choice or a yes or no, stay largely unchanged. MMMU depends on college-level subject knowledge and reasoning over several steps, which short answers about charts and plotting programs do not exercise, and the understanding parameters drift away from it a little more with each round. Only 5\% of each round's \blue{Inspector} data is general visual data, which is too little to hold this ability in place. In our experiments, the share of replay data also proves important for keeping the general abilities of the model.}

\input{tables/public}

%% file: tables/rounds_levels.tex
\begin{table}[t]
\centering
\small
\setlength{\tabcolsep}{4pt}
\caption{\blue{Generation accuracy after each round on BasicChartBench and the chart subset of StructT2IBench. All strategies start from the same model $M_0$. Single diagnosis shares the first round with RSI. Continued training is evaluated once, after 1{,}600 steps on the original data of $M_0$, and is listed under R2. The best strategy in each round is shaded and underlined. R$r$ denotes round $r$, and -- marks a round that is not evaluated.}}
\label{tab:rounds_levels}
\begin{tabular}{@{}lccc|ccc@{}}
\toprule
\multirow{2}{*}{\blue{Strategy}} & \multicolumn{3}{c|}{\blue{BasicChartBench}} & \multicolumn{3}{c}{\blue{StructT2IBench}} \\
\cmidrule(lr){2-4}\cmidrule(l){5-7}
 & \blue{Start} & \blue{R1} & \blue{R2} & \blue{Start} & \blue{R1} & \blue{R2} \\
\midrule
\blue{Continued training} & \blue{26.6} & \blue{--} & \blue{27.3} & \blue{11.0} & \blue{--} & \blue{11.5} \\
\blue{Fixed distribution} & \blue{26.6} & \blue{29.0} & \blue{30.2} & \blue{11.0} & \blue{20.4} & \blue{23.6} \\
\blue{Single diagnosis} & \blue{26.6} & \blue{29.9} & \best{\blue{31.1}} & \blue{11.0} & \blue{20.6} & \blue{24.1} \\
\blue{RSI (ours)} & \blue{26.6} & \best{\blue{29.9}} & \blue{30.8} & \blue{11.0} & \best{\blue{20.6}} & \best{\blue{25.3}} \\
\bottomrule
\end{tabular}
\end{table}

%% file: tables/joint_update.tex
{\color{black}
\begin{table}[t]
\centering
\small
\setlength{\tabcolsep}{4pt}
\caption{Joint and separate updates after the first round. Both use the same data of the first round of RSI. The joint update trains all parameters at once, as in RSI. The separate update trains the generation and the understanding parameters each with the other frozen and merges them afterwards. Code is scored on 232 chart-plotting requests with greedy decoding and at most 16{,}384 new tokens, as the share of programs with valid syntax and the share that execute. The better of the two updates is shaded and underlined.}
\label{tab:joint_update}
\begin{tabular}{@{}llccc@{}}
\toprule
 & Benchmark & Start & Separate & Joint \\
\midrule
\multirow{3}{*}{Painter} & StructT2IBench & 11.0 & 11.7 & \best{20.6} \\
 & BizGenEval & 7.6 & 9.2 & \best{12.5} \\
 & GenEval & 82.9 & 83.3 & \best{84.4} \\
\midrule
\multirow{4}{*}{Inspector} & MME & 2339 & \best{2376} & 2345 \\
 & MMBench & 82.7 & \best{83.6} & 83.5 \\
 & MMMU & 52.1 & \best{50.9} & 50.0 \\
 & MM-Vet & 50.3 & \best{58.2} & 55.5 \\
\midrule
\multirow{2}{*}{Coder} & Valid syntax & 70.7 & 69.0 & \best{71.6} \\
 & Executes & 65.5 & 65.5 & \best{66.0} \\
\bottomrule
\end{tabular}
\end{table}
}

%% file: tables/public.tex
\begin{table}[t]
\centering
\small
\setlength{\tabcolsep}{4pt}
\caption{\blue{General abilities after each round of RSI. GenEval~\citep{ghosh2023geneval} scores the Painter, and MME~\citep{fu2026mme}, MMBench~\citep{liu2024mmbench}, MMMU~\citep{yue2024mmmu}, and MM-Vet~\citep{yu2023mm} the Inspector. Wherever scoring needs a judge model, we use Qwen3.6-35B-A3B. R$r$ denotes round $r$.}}
\label{tab:public}\label{tab:retention}
\begin{tabular}{@{}lcccc@{}}
\toprule
 & Start & R1 & R2 & R3 \\
\midrule
GenEval & 82.9 & \blue{84.4} & \blue{85.2} & \blue{84.6} \\
MME & 2339 & \blue{2345} & \blue{2336} & \blue{2332} \\
MMBench & 82.7 & \blue{83.5} & \blue{83.3} & \blue{83.9} \\
MMMU & 52.1 & \blue{50.0} & \blue{49.8} & \blue{48.9} \\
MM-Vet & \blue{50.3} & \blue{55.5} & \blue{56.5} & \blue{57.1} \\
\bottomrule
\end{tabular}
\end{table}

%% file: sec/5_related_work.tex
\section{Related Work}
\label{sec:related_work}

\textbf{Unified Multimodal Models.} Unified multimodal models provide a shared foundation for visual understanding and generation, spanning different choices of visual representation, modeling objective, and joint training~\citep{xie2025showo,wu2024janus,deng2025bagel}. Beyond architectural unification, interleaved multimodal reasoning connects these capabilities within a task: generated images become intermediate representations that the model can interpret and reason over~\citep{li2025zebracot,qin2026unicot}. This makes the interaction between understanding and generation a central aspect of unified modeling. We examine this interaction across training rounds, where understanding the model's own visual outputs informs the construction of its subsequent training data.
\looseness=-1

\textbf{Recursive Self-Improvement.} Recursive self-improvement couples model updates with the supervision used in later training rounds. Existing approaches recover learning signals from self-generated outputs through outcome verification or learned evaluation~\citep{zelikman2022star,yuan2024selfrewarding}, and extend this process to adaptive task construction when execution supplies verifiable feedback~\citep{zhao2025absolutezero,heng2026eve}. For unified multimodal models, the relationship between understanding and generation provides an additional source of supervision through cross-modal consistency and self-generated visual experience~\citep{hong2025suder,thawkar2026ask,han2026unicorn,hao2026dreaming}. These studies motivate examining how feedback quality evolves alongside the model being trained. Our focus is the role of updated visual understanding in this process: how learning to recognize generation errors changes the examples constructed for subsequent training.
\looseness=-1

\textbf{Synthetic Data Generation.} The value of synthetic training data depends on both the reliability of its supervision and its relevance to the learner. Programmatic visual data construction exposes information about how examples are produced, supporting annotation and verification through rendering code and executable transformations~\citep{yang2025cosyn,heng2026eve}. Adaptive data generation uses learner performance, quality feedback, and coverage gaps to guide the composition of the training set~\citep{khan2025dataenvgym,zhao2024evolvedirector,yan2026dataevolver}. These approaches address complementary aspects of data construction: verifying an example does not by itself establish its usefulness for the current model. Our method combines model-based error diagnosis with programmatic construction, using visual understanding to select training needs and execution records to verify the resulting examples and supervise their visible content.

%% file: sec/6_conclusion.tex
\section{Conclusion}
\label{sec:conclusion}\label{sec:discussion}

We proposed recursive cross-capability self-improvement, a loop in which the text and visual abilities of a UMM produce training data for one another and program execution keeps that data correct. On chart generation, \blue{three rounds raise question accuracy on BasicChartBench from 26.6\% to 31.7\%, while continued training on the original data reaches only 27.3\%.} Verified construction carries \blue{most of} the gain, targeting the model's failures adds \blue{0.6\% on BasicChartBench and 1.7\% on StructT2IBench after two rounds}, and \blue{the Coder and the Inspector} improve along with \blue{the Painter}. These results point toward a broader goal: models whose abilities keep supplying one another with correct training data as they improve.
\looseness=-1

\section*{LLM Usage Statement}
We used Claude (Anthropic) to check grammar and edit text, and to help draw the figures.
The authors checked all content and take full responsibility for it.
\looseness=-1

%% file: appendix/00_implementation_details.tex
\section{Implementation Details}
\label{sec:appendix}

\subsection{Starting Model}
\label{sec:appendix_start}

We start from BAGEL-7B-MoT~\citep{deng2025bagel} and build $M_0$ with a curriculum that goes from easy to hard. \blue{The Coder} is first trained on paired ChartNet code and images with the generation parameters frozen. \blue{The Painter} is then trained with the understanding parameters frozen, first on basic chart types with short text and then on a larger set of programmatically generated charts. We use the checkpoint after 14{,}000 updates of this second stage. Finally, the understanding parameters learn to answer atomic questions about program-rendered charts while keeping \blue{the Coder}, with the generation parameters frozen. This last step starts from the same code-trained checkpoint and changes only the understanding parameters, so combining it with the generation checkpoint equals running it after generation training. The \blue{Inspector} of $M_0$ is \blue{$I_0$}.
\looseness=-1

\subsection{Training in Each Round}

\blue{Each round runs one joint update (\Cref{sec:recursive_update}). It starts from the moving-average weights of the previous round, without the optimizer state, and trains all parameters of the language model, 6.53B on the generation side and 7.63B on the understanding side, while the VAE and the vision encoder stay frozen. The update runs 704 steps on 32 GPUs. Every step takes one batch of generation data and one batch of understanding data, each with 128 examples, so a round sees 90{,}112 samples of each kind, or about 30{,}000 examples three times. The loss adds the text cross-entropy with weight 0.25 to the image generation loss with weight 1. We use AdamW with $\beta=(0.9,0.95)$, a constant learning rate of $2\times10^{-5}$ without warmup \rev{except in the first round, which warms up over 100 steps}, gradient clipping at 1.0, and an exponential moving average with decay 0.995. Images have at most 1{,}536 pixels per side, and the timestep shift is 4. \rev{No condition is dropped for either the generation or the understanding data.}}

\blue{In the generation data,} verified targets of the round make up \rev{70\%} of the examples. Replay of the initial generation data makes up \rev{10\%}, general images \rev{from three captioning sources} 15\%, \rev{with 5\% each,} and editing pairs 5\%.

\blue{In the understanding data,} atomic questions make up \rev{58.3\%}, whole-image error reports \rev{6.7\%}, code 15\%, general visual \rev{question answering 15\%, and text-only data} 5\%. Half of the code examples are the model's own verified programs of the round, weighted by the \blue{Coder} diagnosis. The other half replay the initial code data. Ten percent of the edited-render families are held out in every round to measure the \blue{Inspector}.

\subsection{Construction and Verification}

Each round \rev{selects a fixed 21{,}000 new verified generation targets.} About half of the requests use the curriculum contract wording and the rest use one of six structured forms. The model writes two candidate programs per specification, one greedy and one sampled at temperature 0.6, with at most 6{,}144 new tokens. \blue{The Executor, a sandbox,} runs each program and records every panel. Verification checks the data, text, series and category colors, category order, axes and ticks, grid, value labels, legend, layout, and readability. We keep the first verified candidate per specification and fill each pair of chart type and capability dimension up to its share without borrowing from other pairs. If pairs fall short, we build extra specifications for them in proportion to their observed success rate. Only the first round needed this, with about 30{,}000 extra specifications.

\subsection{Decision Rules}

\Cref{tab:rules} lists the thresholds. All were fixed before the \rev{round in which they were} first \rev{used}.

\input{tables/rules}

%% file: tables/rules.tex
\begin{table}[h]
\centering
\small
\caption{Decision rules of RSI.}
\label{tab:rules}
\begin{tabular}{@{}p{0.34\linewidth}p{0.58\linewidth}@{}}
\toprule
Rule & Setting \\
\midrule
Allocation over pairs & 20\% spread evenly, the rest by violation rate, capped at three times the mean \\
Mastery of a dimension & Satisfaction of at least 0.80 moves the level up, below 0.60 after an advance moves it down \\
Weak attribute value & At least 30 resolved questions, at least 15\% below the best value of the attribute, and below mastery \\
Diagnosis by coverage & Below 0.5\% of weighted training history is introduced, below 5\% is upweighted, otherwise upweighted with extra \blue{Inspector} data. Extra share per value is at most 25\% \\
Wording dependence & A gap of at least 15\% between wordings pairs the value with its weaker wording \\
Curriculum gates & Exact category text of at least 70\% on the current vocabulary and at least 60\% on every earlier vocabulary opens the next vocabulary, exact titles of at least 80\% allow two-word titles, and number format satisfaction of at least 0.80 adds the next format \\
Earlier vocabularies & After a new vocabulary opens, 50\% of specifications keep an earlier one \\
\blue{Coder} diagnosis & At least 60 specifications and at least 10\% below the best value. Extra specifications and code weight grow with the gap, up to three times \\
\bottomrule
\end{tabular}
\end{table}

%% file: appendix/01_request_sets.tex
\section{Request Sets and BasicChartBench}
\label{sec:appendix_sets}

The diagnostic set \blue{has 336 requests}, 48 for each of the seven chart types, and \blue{BasicChartBench has 840 requests over nine chart types}. \blue{In both sets}, the atomic questions are derived from the specification \blue{by a program}. Each question checks one requirement and belongs to one of three groups, structure, text, or binding.

\textbf{Diagnostic set.} The diagnostic set uses the vocabulary of the first curriculum stage, which comes from 12 topics of the generation curriculum with one-word titles and axis titles. Every chart has one panel and three to five categories. For the five chart types with axes, the 48 requests cross the number of series (one, two, or three), the kind of series names (years or words), the size of the values (all below 100, or some at least 100), and the request wording (the curriculum contract or one of six structured forms), with two repeats. Area charts need at least two series, so their one-series cells use two. Pies and rings cross value size and wording with twelve repeats. The first repeat uses integers. The second uses decimals for small values and thousands separators for large values. The set has 9{,}636 questions. \blue{When a new vocabulary opens, the diagnostic set gains a part for it, with the same design and 24 requests per chart type.} This part only feeds the data plan and the vocabulary gate, and it shares no content with any test set.

%

\blue{\textbf{BasicChartBench.} The benchmark has 840 charts with 1{,}960 panels and 52{,}169 questions, between 12 and 171 per chart and 62.1 on average. All panels of a chart have the same type. There are 144 charts each of vertical bars, horizontal bars, lines, and scatter plots, 72 each of stacked areas and histograms, 48 each of pies and rings, and 24 heatmaps, and every type is split evenly over one, two, and four panels. Charts with one, two, and four panels have 7{,}752, 15{,}587, and 28{,}830 questions, or 27.7, 55.7, and 103.0 per chart. The number of categories runs from two to eight, with 118 to 121 charts for each count, and it is spread almost evenly within every chart type and panel count. Category labels have one, two, or three words, with 280 charts each. The two-word labels are written by hand and may contain digits and hyphens, such as ``Q1 2025'' and ``10 am''. Of the 840 charts, 672 have two series and 168 have one, 360 show value labels on the marks, 360 have no grid, 240 a horizontal grid, and 240 a full grid, and half are drawn on a 4:3 canvas ($1280\times960$) and half on a square one ($1280\times1280$). All values are integers, and values and colors are drawn afresh for every chart.}

\blue{\textbf{Answers and scoring.} Every question of BasicChartBench asks for its answer directly. The expected answer is a text string for \rev{26{,}612} questions, a \rev{reading of the marks} for \rev{14{,}421}, an integer for 6{,}688, a set of labels for 2{,}656, an ordered list of labels for 1{,}568, a list of numbers for 168, and a range of numbers for 56. A value read from a mark counts as correct when it lies within 5 units or 10\% of the target. \Cref{tab:bcb_categories} gives the number of questions in each of the 25 categories.}

\textbf{Question coverage.} \rev{Structure questions cover the chart type, the number and layout of the panels, whether a legend and data labels are shown and where the legend sits, the grid, the orientation of the axes, and the geometry and spacing of the marks. Text questions cover the title, the axis titles, the panel letters, the category labels, the legend entries, and the number format. Binding questions cover which value, color, marker, or line style belongs to which category or series, the order of the series, what the data labels show, and how the tick labels are formatted.}

%% file: appendix/02_evaluation_protocol.tex
\section{Evaluation Protocol and External Judge}
\label{sec:appendix_judge}

\textbf{Protocol.} The judge Qwen3.6-35B-A3B answers each atomic question separately with thinking disabled, and its answers are scored by the same deterministic rules as the \blue{Inspector's}. Optical character recognition uses PaddleOCR with rotated reading, so that vertical axis titles are read. For each requested string, we take the best match among the recognized lines. Character similarity is one minus the edit distance to the closest substring, divided by the length of the string.

\blue{\textbf{Reliability of the judge.} All generation results in the main text are scored by the external judge, which shares no weights or training data with our model. To bound the error that the judge itself adds, we also score the reference renders of BasicChartBench, on which every requirement is met. The judge answers 98.3\% of their questions correctly, with 99.0\% on structure, 99.2\% on text, and 97.6\% on binding. Its accuracy stays at or above 97.0\% for every chart type, number of categories, label length, and panel count, and it is lowest on reading a value from a mark, with 96.5\% (\Cref{tab:bcb_categories}).}

%% file: appendix/04_additional_results.tex
\section{Additional Results}
\label{sec:appendix_results}

\begin{figure}[t]
\centering
\includegraphics[width=\linewidth]{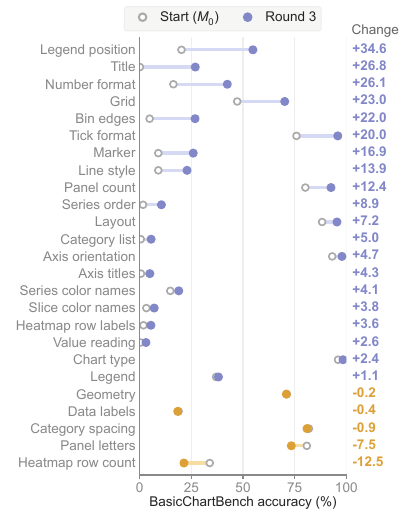}
\caption{\blue{Accuracy of the Painter on BasicChartBench for each question category, at the start and after round 3, ordered by the change.}}
\label{fig:bcb_categories}
\end{figure}

\blue{\Cref{tab:understanding} gives the results of the Inspector on the fixed chart understanding set and on renders with controlled errors for RSI and the controls, and \Cref{tab:coding} gives those of the Coder.} \blue{\Cref{tab:bcb_categories} gives the accuracy on BasicChartBench for each question category after each round, and \Cref{fig:bcb_categories} compares the start with round 3.}

\input{tables/understanding}

\input{tables/coding}

\input{tables/bcb_categories}

%% file: tables/understanding.tex
\begin{table}[t]
\centering
\small
\setlength{\tabcolsep}{2.5pt}
\caption{\blue{Accuracy (\%) of the Inspector for RSI and the controls. The fixed chart understanding set has 342 charts with 3{,}939 question-answering questions (\emph{All}), which are either open or yes-or-no questions, and 1{,}026 verification questions (\emph{Verif.}). The renders with controlled errors have 17{,}946 questions (\emph{All}), of which 1{,}030 ask about the property that the error changes (\emph{Changed}). Only RSI is measured on these renders. Continued training updates only the Painter, so its Inspector is that of the start. Fixed distr.\ and single diagn.\ are the fixed distribution and single diagnosis. R$r$ denotes round $r$, and the best value in each column is shaded and underlined.}}
\label{tab:understanding}
\begin{tabular}{@{}lcccc|cc@{}}
\toprule
\multirow{2}{*}{\blue{Model}} & \multicolumn{4}{c|}{\blue{Understanding set}} & \multicolumn{2}{c}{\blue{Controlled errors}} \\
\cmidrule(lr){2-5}\cmidrule(l){6-7}
 & \blue{All} & \blue{Open} & \blue{Yes/no} & \blue{Verif.} & \blue{All} & \blue{Changed} \\
\midrule
\blue{Start ($M_0$)} & \blue{72.7} & \blue{69.2} & \blue{76.3} & \blue{17.0} & \blue{66.4} & \blue{60.0} \\
\midrule
\blue{Fixed distr., R1} & \blue{92.2} & \blue{90.1} & \blue{94.5} & \blue{78.6} & \blue{--} & \blue{--} \\
\blue{Fixed distr., R2} & \blue{93.2} & \blue{91.1} & \blue{95.7} & \best{\blue{84.3}} & \blue{--} & \blue{--} \\
\blue{Single diagn., R2} & \blue{92.7} & \blue{90.8} & \blue{94.9} & \blue{82.7} & \blue{--} & \blue{--} \\
\midrule
\blue{RSI, R1} & \blue{93.0} & \blue{91.3} & \blue{95.1} & \blue{80.4} & \blue{88.2} & \blue{96.6} \\
\blue{RSI, R2} & \blue{89.9} & \blue{91.6} & \blue{88.3} & \blue{75.1} & \blue{83.2} & \best{\blue{98.0}} \\
\blue{RSI, R3} & \best{\blue{94.2}} & \best{\blue{92.2}} & \best{\blue{96.5}} & \blue{84.1} & \best{\blue{89.8}} & \blue{95.2} \\
\bottomrule
\end{tabular}
\end{table}

%% file: tables/coding.tex
\begin{table}[t]
\centering
\small
\setlength{\tabcolsep}{4pt}
\caption{\blue{The Coder for RSI and the controls on 232 chart-plotting requests, with greedy decoding and at most 16{,}384 new tokens. \emph{Valid syntax} and \emph{Executes} are the shares (\%) of programs that parse and that run, and \emph{Token limit} is the share that reaches the limit of new tokens. Single diagnosis shares its first round with RSI. R$r$ denotes round $r$, and the best value in each column is shaded and underlined.}}
\label{tab:coding}
\begin{tabular}{@{}lccc@{}}
\toprule
\blue{Model} & \blue{Valid syntax} & \blue{Executes} & \blue{Token limit} \\
\midrule
\blue{Start ($M_0$)} & \blue{70.7} & \blue{65.5} & \blue{29.3} \\
\midrule
\blue{Fixed distribution, R1} & \blue{69.8} & \blue{65.1} & \blue{30.2} \\
\blue{Fixed distribution, R2} & \blue{67.7} & \blue{60.8} & \blue{31.0} \\
\blue{Single diagnosis, R2} & \blue{66.8} & \blue{61.2} & \blue{31.9} \\
\midrule
\blue{RSI, R1} & \blue{71.6} & \best{\blue{66.0}} & \blue{28.0} \\
\blue{RSI, R2} & \blue{70.3} & \blue{65.1} & \blue{30.2} \\
\blue{RSI, R3} & \best{\blue{73.7}} & \best{\blue{66.0}} & \best{\blue{26.3}} \\
\bottomrule
\end{tabular}
\end{table}

%% file: tables/bcb_categories.tex
\begin{table}[t]
\centering
\small
\setlength{\tabcolsep}{3.5pt}
\caption{\blue{Question accuracy (\%) of the Painter on BasicChartBench for each of the 25 question categories, ordered by the number of questions. R$r$ denotes round $r$, and Ref.\ is the accuracy of the external judge on the reference renders.}}
\label{tab:bcb_categories}
\begin{tabular}{@{}lrccccc@{}}
\toprule
\blue{Category} & \blue{Questions} & \blue{Start} & \blue{R1} & \blue{R2} & \blue{R3} & \blue{Ref.} \\
\midrule
\blue{Value reading} & \blue{14{,}421} & \blue{0.4} & \blue{2.1} & \blue{2.9} & \blue{3.0} & \blue{96.5} \\
\blue{Data labels} & \blue{5{,}712} & \blue{18.8} & \blue{16.0} & \blue{16.5} & \blue{18.4} & \blue{98.1} \\
\blue{Geometry} & \blue{3{,}864} & \blue{71.2} & \blue{71.8} & \blue{72.1} & \blue{71.0} & \blue{98.5} \\
\blue{Axis titles} & \blue{3{,}528} & \blue{0.6} & \blue{1.6} & \blue{3.9} & \blue{4.9} & \blue{99.1} \\
\blue{Legend} & \blue{3{,}296} & \blue{37.0} & \blue{32.9} & \blue{36.8} & \blue{38.1} & \blue{97.7} \\
\blue{Series color names} & \blue{3{,}192} & \blue{14.8} & \blue{19.2} & \blue{18.9} & \blue{18.9} & \blue{98.6} \\
\blue{Tick format} & \blue{2{,}016} & \blue{75.9} & \blue{93.7} & \blue{95.8} & \blue{95.9} & \blue{100.0} \\
\blue{Title} & \blue{1{,}960} & \blue{0.1} & \blue{15.4} & \blue{21.2} & \blue{26.9} & \blue{99.9} \\
\blue{Chart type} & \blue{1{,}792} & \blue{96.0} & \blue{98.6} & \blue{97.8} & \blue{98.4} & \blue{100.0} \\
\blue{Panel letters} & \blue{1{,}680} & \blue{80.9} & \blue{70.0} & \blue{65.9} & \blue{73.4} & \blue{99.9} \\
\blue{Marker} & \blue{1{,}344} & \blue{9.0} & \blue{23.6} & \blue{25.8} & \blue{25.9} & \blue{99.6} \\
\blue{Series order} & \blue{1{,}344} & \blue{1.6} & \blue{6.4} & \blue{9.2} & \blue{10.5} & \blue{97.4} \\
\blue{Category spacing} & \blue{1{,}176} & \blue{81.9} & \blue{84.5} & \blue{85.9} & \blue{81.0} & \blue{99.4} \\
\blue{Slice color names} & \blue{1{,}121} & \blue{3.2} & \blue{5.3} & \blue{4.8} & \blue{7.0} & \blue{99.6} \\
\blue{Category list} & \blue{1{,}120} & \blue{0.5} & \blue{2.5} & \blue{4.1} & \blue{5.5} & \blue{99.6} \\
\blue{Layout} & \blue{1{,}120} & \blue{88.3} & \blue{96.5} & \blue{89.7} & \blue{95.5} & \blue{100.0} \\
\blue{Number format} & \blue{840} & \blue{16.3} & \blue{50.9} & \blue{53.8} & \blue{42.4} & \blue{99.9} \\
\blue{Panel count} & \blue{840} & \blue{80.2} & \blue{94.3} & \blue{85.0} & \blue{92.6} & \blue{100.0} \\
\blue{Grid} & \blue{627} & \blue{47.2} & \blue{62.8} & \blue{69.9} & \blue{70.2} & \blue{97.8} \\
\blue{Axis orientation} & \blue{336} & \blue{93.2} & \blue{92.9} & \blue{96.1} & \blue{97.9} & \blue{100.0} \\
\blue{Line style} & \blue{288} & \blue{9.0} & \blue{20.1} & \blue{19.4} & \blue{22.9} & \blue{98.6} \\
\blue{Legend position} & \blue{272} & \blue{20.2} & \blue{37.5} & \blue{51.8} & \blue{54.8} & \blue{100.0} \\
\blue{Bin edges} & \blue{168} & \blue{4.8} & \blue{13.7} & \blue{21.4} & \blue{26.8} & \blue{100.0} \\
\blue{Heatmap row count} & \blue{56} & \blue{33.9} & \blue{32.1} & \blue{23.2} & \blue{21.4} & \blue{100.0} \\
\blue{Heatmap row labels} & \blue{56} & \blue{1.8} & \blue{5.4} & \blue{1.8} & \blue{5.4} & \blue{100.0} \\
\midrule
\blue{All} & \blue{52{,}169} & \blue{26.6} & \blue{29.9} & \blue{30.8} & \blue{31.7} & \blue{98.3} \\
\bottomrule
\end{tabular}
\end{table}

%% file: appendix/05_data_distribution_changes.tex
\section{Data Distribution Changes}
\label{sec:appendix_changes}

The rules above were fixed before the round in which they were first used. We list every change that affects the reported rounds, when it was made, and why.

\textbf{Before round 1.} Three fixes were made while the data of round 1 were built, before any training. The pairing of number formats with prompt formats was changed from a fixed probability to a rate that divides the target share by the frequency of the value, because the fixed rate sent 92\% of the first-round prompts to one format. The overall share of prompts in the mixed format was then held between 40\% and 60\%. The content checker was fixed for line charts whose programs plot category names directly, which it had rejected almost always, and the saved traces of round 1 were checked again without running the programs again.